\documentclass[11pt, a4paper]{article}

\usepackage[margin=1in]{geometry}
\usepackage[utf8]{inputenc}
\usepackage[T1]{fontenc}
\usepackage{lmodern}

\usepackage{hyperref}
\usepackage{url}
\usepackage{booktabs}
\usepackage{amsfonts}
\usepackage{amsmath}
\usepackage{nicefrac}
\usepackage{microtype}
\usepackage{fancyhdr}
\usepackage{graphicx}
\usepackage{cleveref} 
\usepackage{authblk}  

\graphicspath{{./}}       

\title{\textbf{KAN-FPN-Stem: A KAN-Enhanced Feature Pyramid Stem for Boosting ViT-based Pose Estimation}}

\author[1]{\textbf{Haonan Tang}}
\affil[1]{WuHan University of Technology, WuHan, HuBei, China}
\affil[ ]{\texttt{lancera.thn@gmail.com}}

\date{}

\begin{document}
\maketitle

\begin{abstract}
Vision Transformers (ViT) have demonstrated significant promise in dense prediction tasks such as pose estimation. However, their performance is frequently constrained by the overly simplistic front-end designs employed in models like ViTPose. This naive patchification mechanism struggles to effectively handle multi-scale variations and results in irreversible information loss during the initial feature extraction phase. To overcome this limitation, we introduce a novel KAN-enhanced FPN-Stem architecture. Through rigorous ablation studies, we first identified that the true bottleneck for performance improvement lies not in plug-and-play attention modules (e.g., CBAM), but in the post-fusion non-linear smoothing step within the FPN. Guided by this insight, our core innovation is to retain the classic "upsample-and-add" fusion stream of the FPN, but replace its terminal, standard linear 3x3 smoothing convolution with a powerful KAN-based convolutional layer. Leveraging its superior non-linear modeling capabilities, this KAN-based layer adaptively learns and rectifies the "artifacts" generated during the multi-scale fusion process. Extensive experiments on the COCO dataset demonstrate that our KAN-FPN-Stem achieves a significant performance boost of up to +2.0 AP over the lightweight ViTPose-S baseline. This work not only delivers a plug-and-play, high-performance module but, more importantly, reveals that: the performance bottleneck in ViT front-end often lies not in 'feature refinement' (Attention), but in the quality of 'feature fusion' (Fusion). Furthermore, it provides an effective path to address this bottleneck through the introduction of the KAN operator.
\end{abstract}

\vspace{1em}

\section{Introduction}
Vision Transformers (ViT) \cite{xiao2021early} are profoundly reshaping the field of computer vision with their powerful global modeling capabilities. In human pose estimation, a classic dense prediction task, plain ViT models exemplified by ViTPose \cite{xu2022vitpose} have demonstrated that highly competitive performance can be achieved even without relying on complex convolutional network designs. This has established pure-Transformer-based, minimalist architectures as a research direction of great interest.

However, a performance bottleneck lurks beneath this design simplicity. Models like ViTPose adopt the native patch embedding mechanism from ViT, which is essentially a large-stride, non-overlapping convolution. This ``single-step'' feature extraction approach runs counter to the ``hierarchical'' design philosophy of modern convolution networks, inevitably causing significant loss of fine-grained local features crucial for pose recognition at the very early stages. Furthermore, its single embedding scale makes it difficult to effectively cope with the ubiquitous human scale variations in images.

To overcome this bottleneck and fully unleash the potential of small ViT models on pose estimation tasks, we undertook a series of progressive explorations. We first effectively addressed the native feature extraction and multi-scale adaptability issues by introducing a Stem module based on CNNs and the Feature Pyramid Network (FPN) \cite{lin2017feature}. However, as we further attempted to enhance features using attention mechanisms (e.g., CBAM \cite{woo2018cbam}), we arrived at a critical insight: simply stacking attention modules within the network did not yield significant performance gains. This led us to realize that the true bottleneck might not be that the features themselves were insufficiently ``pure,'' but rather that ``the feature fusion operation in FPN was overly simplistic.''

Building on this profound insight, we did not opt for a `complete overhaul' of the FPN. Instead, we propose a more sophisticated solution: our KAN-FPN-Stem. We retain the FPN's classic `upsample + element-wise addition' fusion stream, as our ablation studies (shown in Sec. 4.2.3) confirmed its efficacy in preserving multi-scale information. Our core innovation lies in precisely targeting this ``overly simplistic'' bottleneck: we replace the FPN's terminal, standard linear 3x3 smoothing convolution with a powerful KAGN convolutional layer. Leveraging its superior non-linear modeling capabilities, this KAGN layer adaptively learns and rectifies the ``artifacts'' generated during the multi-scale fusion process, thereby fundamentally improving the fusion quality without compromising the FPN's efficient structure.

We conducted extensive experiments on the public MS COCO dataset. Our ablation studies clearly illustrate the logical progression from a simple CNN Stem to our final KAN-FPN-Stem solution. Ultimately, our optimal model achieves a significant performance boost of up to 2.0 AP over our reproduced ViTPose-S baseline. Our main contributions are as follows:
\begin{itemize}
  \item Through a series of rigorous ablation studies, we reveal that in ViT frontend design, the ``quality'' of feature fusion is far more critical than simple feature ``enhancement.''
  \item We propose the KAN-FPN-Stem, a novel hybrid architecture that replaces the FPN's terminal smoothing convolution with a KAGN convolution, offering a more powerful solution for mitigating multi-scale fusion artifacts.
  \item Our final model achieves a +2.0 AP gain on the lightweight ViTPose-S, validating our hypothesis that the quality of multi-scale feature fusion is the true bottleneck in our task, and providing new insights for designing more efficient CNN-ViT hybrid networks.
\end{itemize}

\section{Related Work}

\subsection{Encoder-Decoder based Heatmap Pose Estimation}
The dominant paradigm in modern human pose estimation is the Encoder-Decoder architecture for heatmap regression, an idea crystallized in the seminal work of Xiao et al., Simple Baselines \cite{xiao2018simple}. In this framework, a deep convolutional network (the encoder) first downsamples the image into high-level features. Subsequently, a simple deconvolutional decoder upsamples these features to generate a heatmap corresponding to each keypoint. This simple yet effective strategy of ``encoding features-decoding heatmaps'' has become the established methodology in the field, and a vast body of subsequent research, including our own, builds upon this paradigm.

\subsection{Vision Transformers in Pose Estimation}
In recent years, the Vision Transformer (ViT) \cite{xiao2021early} has introduced a new paradigm to the computer vision landscape with its powerful global dependency modeling. In the realm of pose estimation, ViTPose \cite{xu2022vitpose}, proposed by Xu et al., successfully demonstrated that a plain, non-hierarchical ViT encoder can serve as a powerful feature extractor, even surpassing meticulously designed convolutional networks. The success of ViTPose highlights the unique advantage of the ViT architecture in capturing long-range dependencies between human body parts.

However, this design simplicity also inherits the fundamental flaws of the native ViT in frontend feature processing. Its ``single-step'' patch embedding mechanism is essentially a large-stride, non-overlapping convolution, which runs counter to the validated, progressive feature extraction paradigm of modern CNNs \cite{xiao2021early}. This design inevitably discards a significant amount of local texture and edge information --- crucial for precise keypoint localization --- at the very initial stages of feature extraction. It also limits the model's flexible adaptability to multi-scale targets within the image.

\subsection{Hybrid CNN-ViT Architectures}
To address the deficiencies of native ViT in frontend feature extraction, a prominent research direction is the construction of hybrid CNN-ViT architectures. The work by Xiao et al. \cite{xiao2021early} is a typical representative of this approach. They demonstrated through experiments that replacing the native Patch Embedding with a CNN Stem, composed of a few standard convolutions, can significantly improve ViT's training stability and final performance. This line of work confirms that introducing the local inductive bias of convolutions in the early stages of ViT is crucial for the model to effectively learn low-level visual features.

However, the main contribution of these hybrid architectures lies in optimizing the ViT's input quality; their output is still, in essence, a single-scale feature map. For a complex task like human pose estimation, which requires simultaneous perception of large, nearby targets and small, distant ones, relying solely on a single-scale, high-level feature map makes it difficult to account for the details and contours of different-sized human bodies. This has become a new bottleneck limiting further performance improvements.

\subsection{Multi-Scale Feature Fusion}
To solve the aforementioned single-scale feature limitation, effectively handling multi-scale variations in targets is key to success in dense prediction tasks. The Feature Pyramid Network (FPN) \cite{lin2017feature}, proposed by Lin et al., provides a classic and efficient solution. FPN utilizes a top-down pathway to fuse high-level, semantically-strong feature maps with high-resolution, semantically-weaker feature maps from the bottom-up pathway via upsampling. This design ensures that every level of the final output pyramid is rich in both semantic information and spatial detail, greatly enhancing the model's multi-scale perception capabilities.

However, in FPN and its numerous successors (e.g., PANet \cite{liu2018path}, BiFPN \cite{tan2020efficientdet}), the core fusion operation is typically a simple element-wise addition. This linear superposition, while simple and efficient, implicitly assumes that features from different levels can be fused equally and without conflict. Its modeling capacity is limited when dealing with aliasing artifacts produced by upsampling or when fusing semantically disparate features. This fixed fusion strategy leaves ample room for exploration into more intelligent and powerful non-linear fusion paradigms.

\subsection{Attention Mechanisms in Vision}
Attention mechanisms have been widely proven as an effective means of enhancing CNN performance. Modules such as CBAM \cite{woo2018cbam}, with their lightweight designs, enable the network to adaptively learn ``what to attend to'' (channel) and ``where to attend to'' (spatial), thereby enhancing useful features and suppressing irrelevant information. These modules are often regarded as effective ``plug-and-play'' components that can reliably boost performance across various vision tasks.

However, the effectiveness of such ``plug-and-play'' modules is not absolute; it is highly dependent on the task and deployment strategy. Simple attention enhancement does not necessarily solve the more fundamental, structural problems in multi-scale feature fusion. This observation inspired us that the real bottleneck may not be the ``purity'' of the features, but that the fusion operation itself is too simplistic. Therefore, a new method capable of fundamentally restructuring the ``feature fusion'' process is required.

\subsection{Kolmogorov-Arnold Networks}
Recently, Kolmogorov-Arnold Networks (KAN) \cite{liu2024kan}, proposed by Liu et al.\@, have offered a completely new perspective on neural network design. Unlike traditional Multi-Layer Perceptrons (MLPs), which interleave linear transformations with fixed non-linear activation functions (e.g., ReLU), KAN places learnable, univariate functions --- often based on splines or polynomials --- directly onto the edges of the network. This change allows the non-linear transformations themselves to become learnable objects, gifting the model with unparalleled function-fitting capabilities.

This concept was quickly extended to convolutional networks, spawning work such as KAN-based convolutions \cite{liang2024kolmogorov}, which replace the fixed activation functions in traditional convolutional layers with a set of learnable non-linear functions. This revolutionary property provides us with the perfect tool to solve the fusion limitations of FPN. As we argued previously, FPN's linear addition is limited in handling complex fusion artifacts, and our exploration of attention mechanisms also indicated the need for a more fundamental approach to restructuring the fusion operation. A KAN-based convolutional layer, by virtue of its powerful non-linear modeling capabilities, can theoretically learn an optimal, complex non-linear function specifically for feature fusion, thereby achieving a far more intelligent and efficient fusion than simple addition. Our work represents the first exploration and successful validation of this idea in the domain of pose estimation.

\section{Method}

\subsection{FPN-Stem Foundation}
\begin{figure}[htbp]
  \centering
  \includegraphics[width=0.9\linewidth]{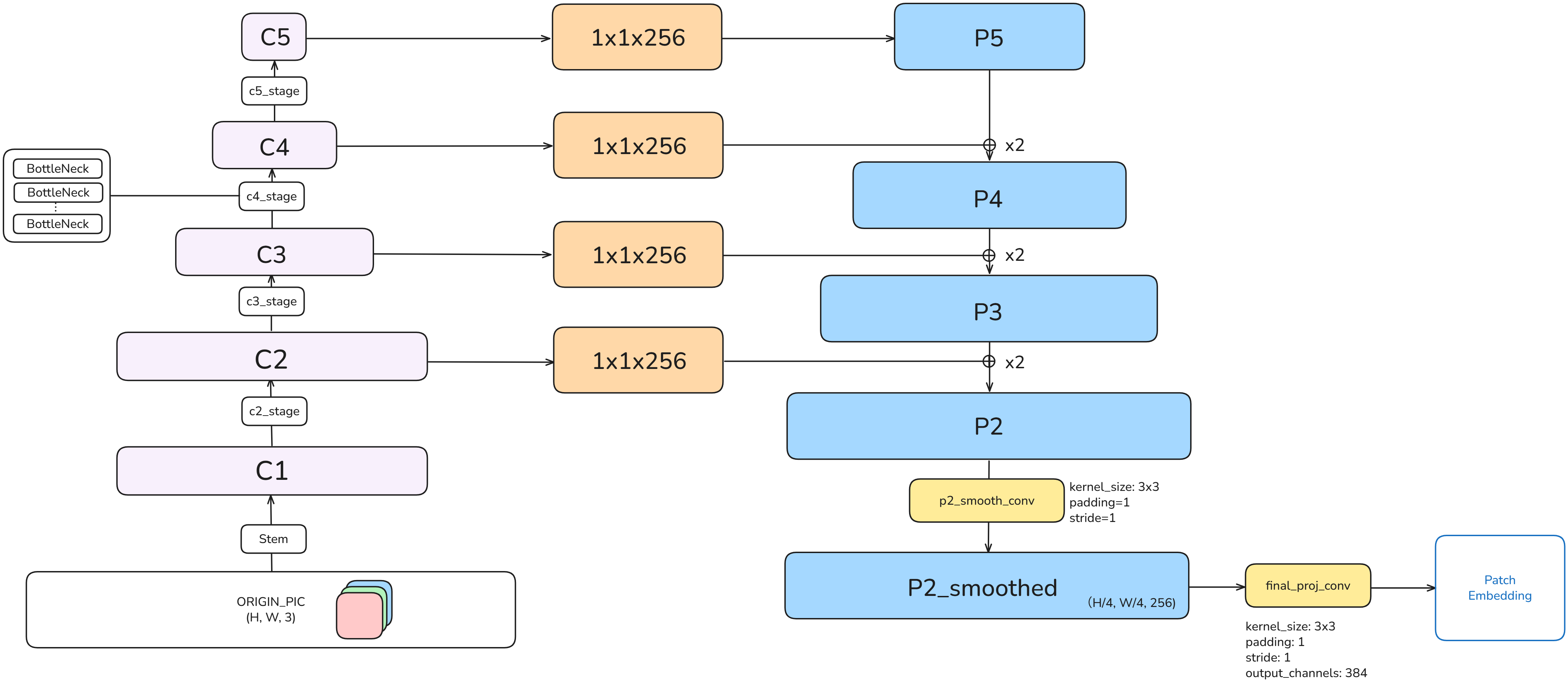} 
  \caption{The architecture of our FPN-Stem.}
  \label{fig:fpn_stem}
\end{figure}

The architecture of our foundational FPN-Stem is illustrated in \cref{fig:fpn_stem}. While experiments have shown that a CNN Stem can mitigate the low-level feature loss caused by ViTPose's native Patch Embedding, its own multi-scale representation capability remains limited. To address both issues concurrently, our method is built upon this more powerful FPNStem module. This module is designed to resolve both the feature loss from the coarse-grained patchification of the native Patch Embedding and the single-scale limitation of a simple CNN Stem.

\begin{itemize}
    \item \textbf{Architecture Upgrade:} The FPNStem integrates a complete ResNet-50 backbone, which generates a feature pyramid comprising four distinct scales (C2, C3, C4, C5) in its bottom-up pathway.
    \item \textbf{Fusion Mechanism:} The module then employs a classic top-down pathway and lateral connections to fuse high-level, semantically-strong features (like C5) with low-level, fine-grained spatial features (like C2). This process ensures that even the highest-resolution feature map is enriched with semantic information.
    \item \textbf{Output Method:} Crucially, the FPNStem exclusively selects the post-fusion p2\_out feature map, which possesses the highest resolution and is the most information-dense. This map is then projected through a 1x1 convolution to the ViT's required dimensionality and fed as a single input to the subsequent Transformer modules.
\end{itemize}

We replace the original shallow CNN Stem with a complete ResNet-50 equipped with an FPN fusion mechanism. Consequently, the input fed to the ViT is no longer merely simple low-level features, but rather a high-quality single feature map, deeply extracted and fully enriched with fused multi-scale information.

\subsection{Proposed Enhancements}
\begin{figure}[htbp]
  \centering
  \includegraphics[width=0.9\linewidth]{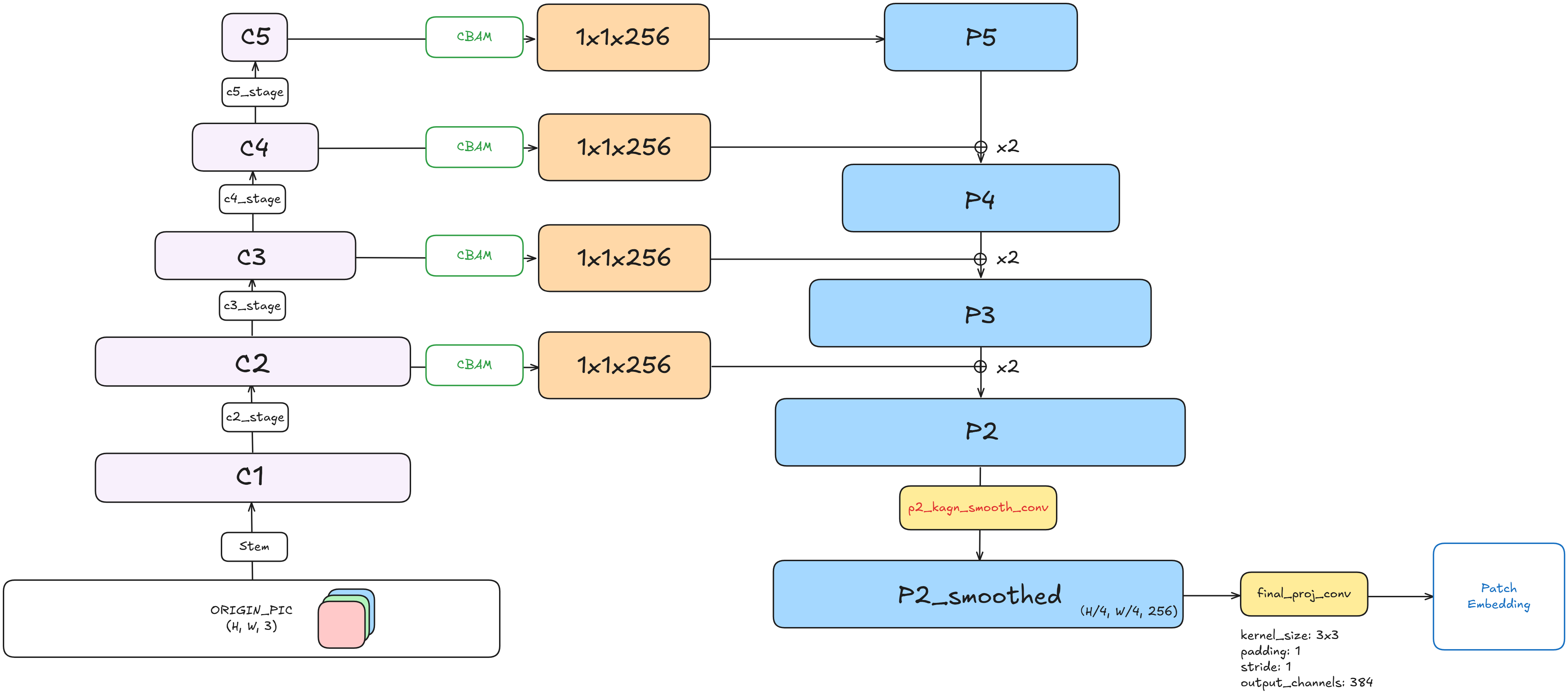} 
  \caption{The architecture of our KAGN-FPN-Stem.}
  \label{fig:kagn_stem}
\end{figure}

Our final proposed model, depicted in \cref{fig:kagn_stem}, is built upon the FPNStem foundation, synergistically integrating a strategic attention mechanism and non-linear smoothing to form an advanced frontend feature extractor.

\begin{itemize}
    \item \textbf{Main Framework:} The model retains the core ResNet-50 + FPN architecture for generating and preliminarily fusing multi-scale features.
    \item \textbf{Core Change 1 (Lateral CBAM):} Inspired by the ``feature selection'' concept in HS-FPN \cite{shi2024hsfpn}, we pre-pended a CBAM module to each lateral connection path of the FPN.
    \item \textbf{Core Change 2 (KAGN Smooth Conv):} We replaced the standard 3x3 smoothing convolution --- used post-fusion at the highest resolution level (P2) in the FPN --- with a BottleNeckKAGNConv2DLayer. This step aims to leverage the powerful non-linear modeling capabilities of KAN to eliminate the fusion ``artifacts'' arising from upsampling and feature addition.
\end{itemize}

\section{Experiments}

\subsection{Implementation Details}
All our experiments were conducted on the MS COCO 2017 dataset. We followed the standard training pipeline of ViTPose. Our baseline model is ViTPose-S. All models were trained for 210 epochs on a single 8GB 4070 GPU. The base learning rate was set to 1e-4, preceded by a 500-iteration linear warmup, and we employed a MultiStepLR decay strategy (decaying at the 170th and 200th epochs). We used a batch size of 16.

\subsection{Ablation Study}
\begin{table}[ht]
  \centering
  \caption{Ablation study on the effectiveness of our components. All models are built on the ViTPose-S baseline.}
  \label{tab:ablation}
  \begin{tabular}{l c c}
    \toprule
    Method & AP (\%) & $\Delta$ AP \\
    \midrule
    Baseline (ViTPose-S, Stage 0) & 72.5 & - \\
    + CNN Stem (Stage 1) & 73.3 & +0.8 \\
    + FPN-Stem (Stage 2) & 74.0 & +1.5 \\
    \midrule
    + FPN-Stem + CBAM (Backbone, Stage 3) & 74.1 & +1.6 \\
    + FPN-Stem + Lateral CBAM (Stage 5) & 73.8 & +1.3 \\
    \midrule
    + KAGN-Fuse (Stage 6) & 74.3 & +1.8 \\
    \textbf{Ours (Stage 4: Lat. CBAM + KAGN Smooth)} & \textbf{74.5} & \textbf{+2.0} \\
    \bottomrule
  \end{tabular}
\end{table}

\subsubsection{The Effectiveness of FPN-Stem}
\begin{figure}[htbp]
  \centering
  \includegraphics[width=0.8\linewidth]{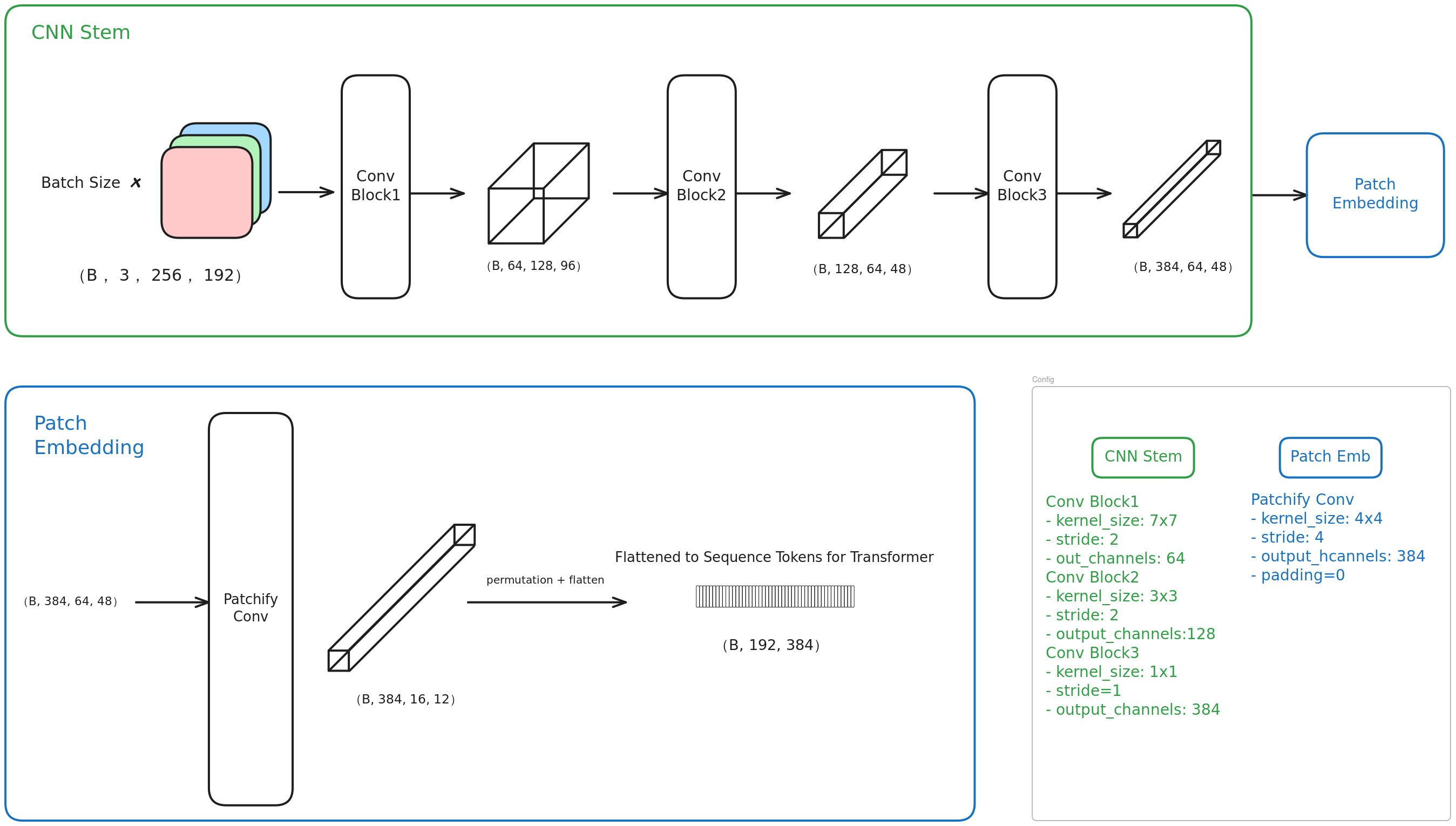} 
  \caption{The architecture of our CNN-Stem.}
  \label{fig:cnn_stem}
\end{figure}

We first conducted a series of experiments to validate the effectiveness of the FPN-Stem in addressing the multi-scale problem. As shown in \cref{tab:ablation}, our reproduced ViTPose-S (Baseline) achieved 72.5 AP. Subsequently, we replaced the native Patch Embedding with a progressive CNN Stem (illustrated in \cref{fig:cnn_stem}). The final performance increased to 73.3 AP (+0.8 AP).

\subsubsection{Exploration of Attention Mechanisms}
\begin{figure}[htbp]
  \centering
  \includegraphics[width=0.8\linewidth]{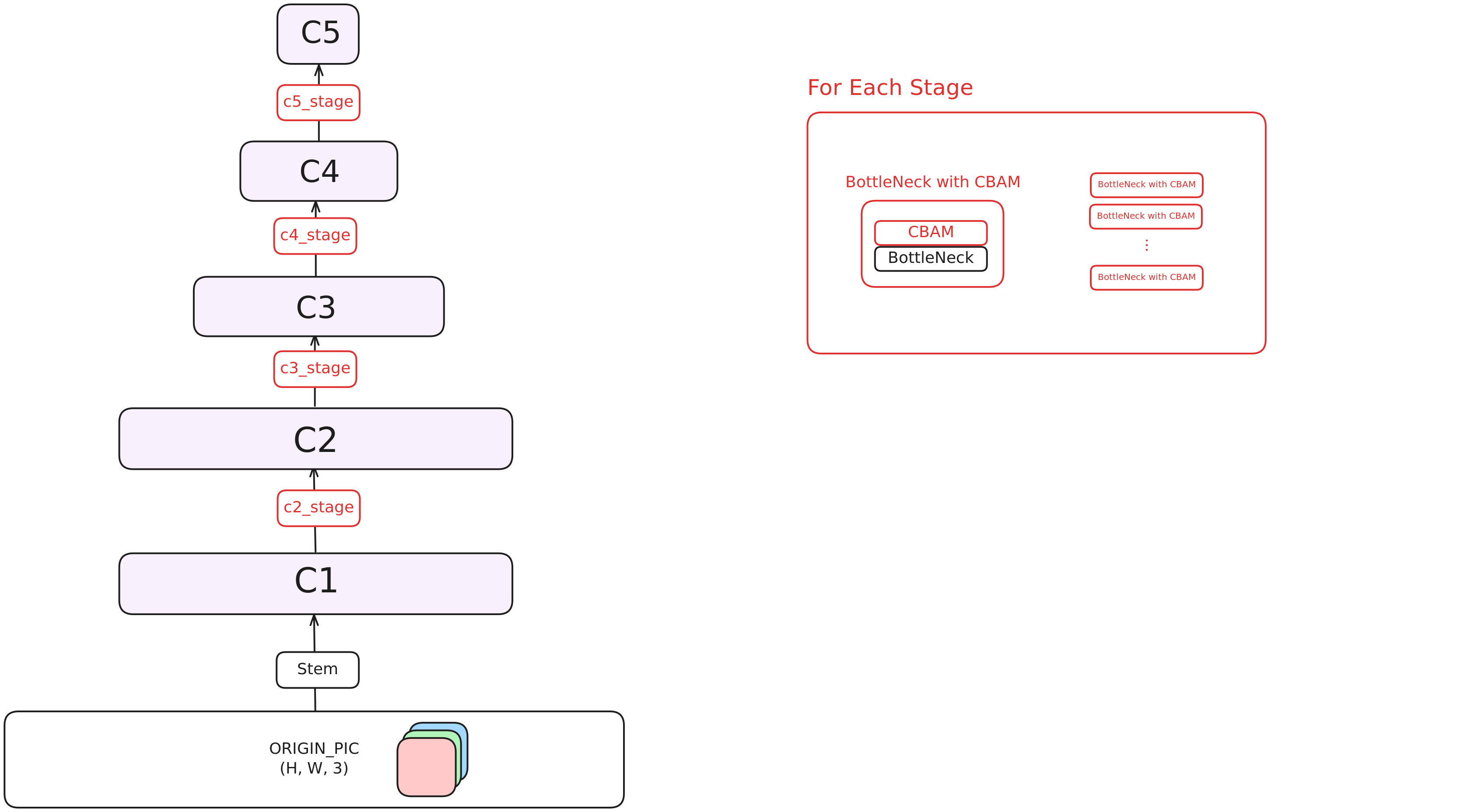} 
  \caption{Illustration of the `carpet-style' attention strategy (Phase 3).}
  \label{fig:add_cbam}
\end{figure}

\begin{figure}[htbp]
  \centering
  \includegraphics[width=0.8\linewidth]{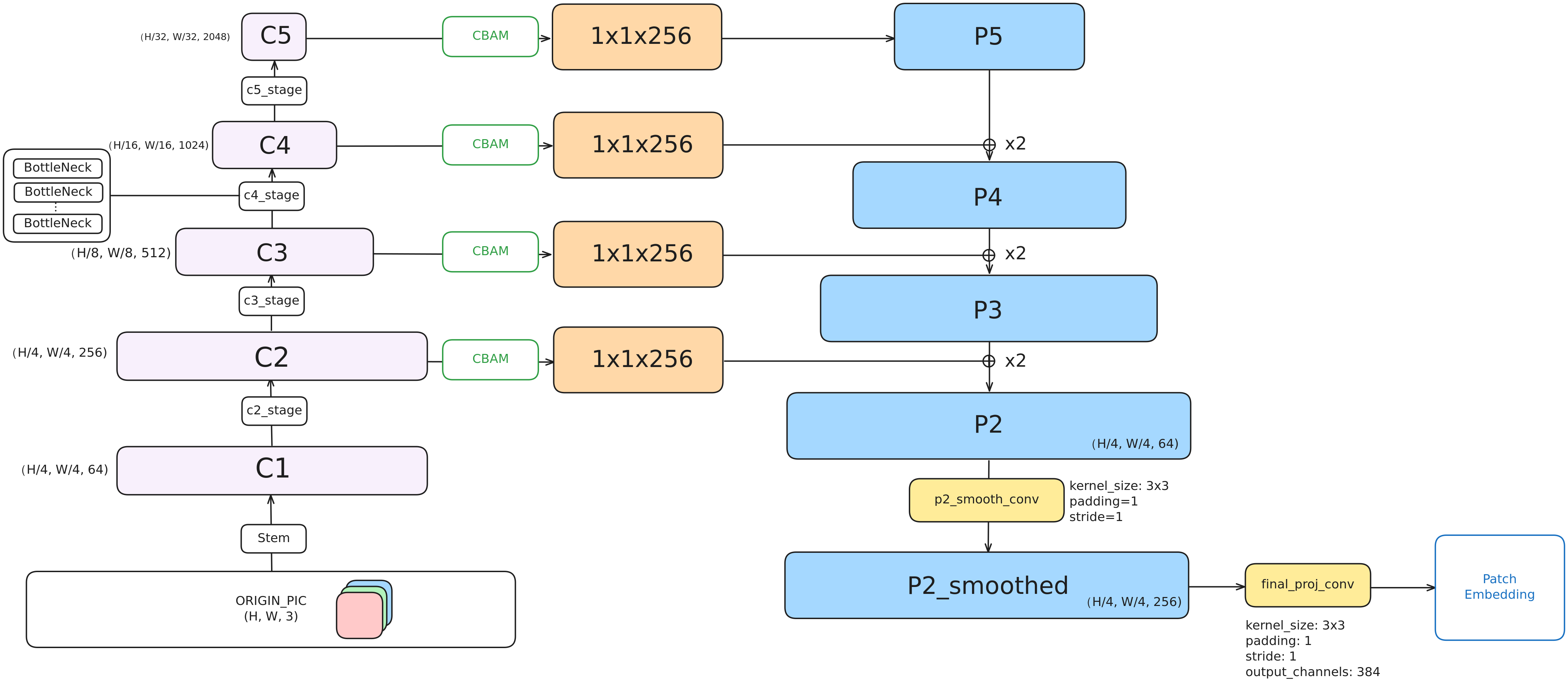} 
  \caption{Illustration of the Lateral CBAM strategy (Phase 5).}
  \label{fig:cbam_mimic}
\end{figure}
Building upon the FPN-Stem, we explored the effectiveness of attention mechanisms. First, we attempted a `blanket' strategy, inserting CBAM modules into every block of the backbone (Stage 3) (as shown in \cref{fig:add_cbam}). The results showed this approach yielded only a marginal improvement of 0.1 AP. Second, inspired by HS-FPN, we tried a more sophisticated strategy: applying CBAM only to the FPN's lateral connections (Stage 5) (as shown in \cref{fig:cbam_mimic}). However, the performance dropped to 73.8 AP.

\subsubsection{KAGN Smoothing as Core Drivers}
\begin{figure}[htbp]
  \centering
  \includegraphics[width=0.8\linewidth]{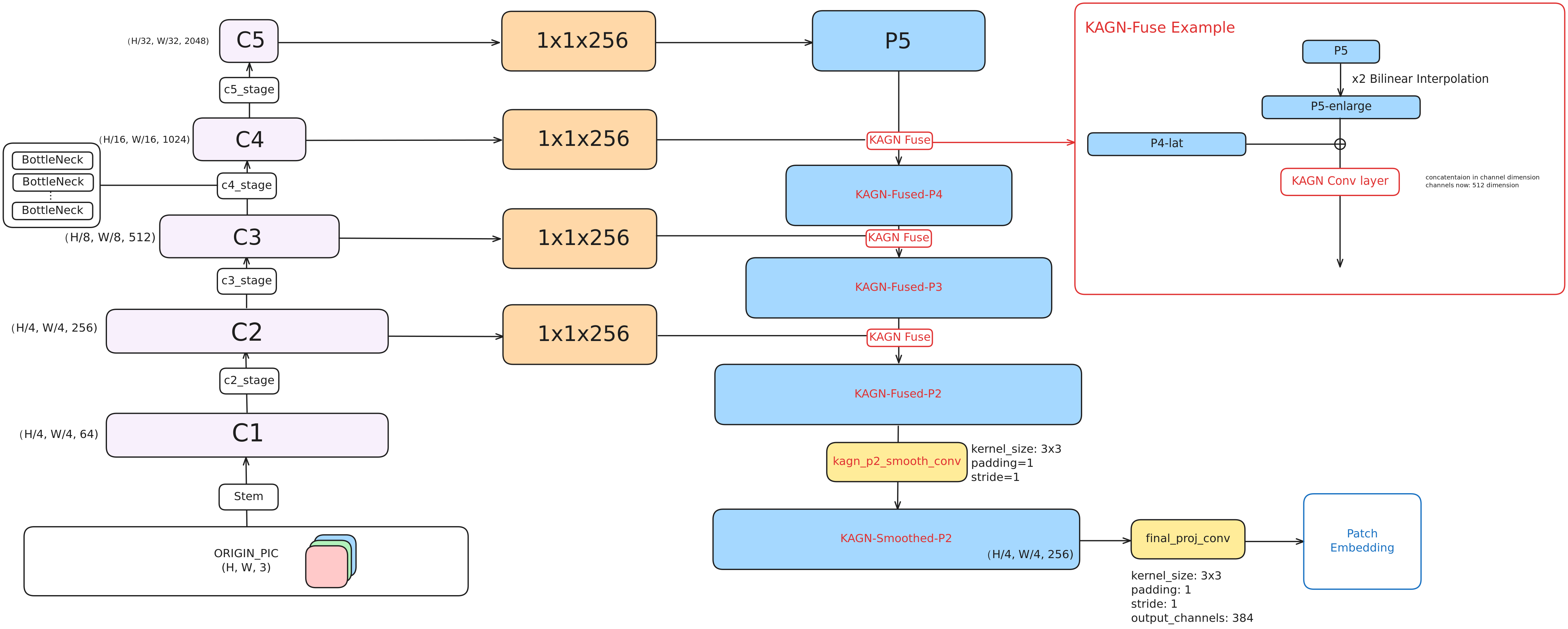} 
  \caption{The architecture of the KAGN-Fuse paradigm (Phase 6).}
  \label{fig:kagn_fuse}
\end{figure}
In the previous subsection, we found that simple attention strategies, such as Lateral CBAM (Stage 5), failed to improve performance. This finding shifted our research focus from `attention' to the `FPN fusion operation itself'. Therefore, building on the Stage 5 model (73.8 AP), we introduced the KAGN Smooth Conv, which constitutes our final model (architecture shown in \cref{fig:kagn_stem}).

The result was significant: As shown in \cref{tab:ablation}, model performance surged from 73.8 AP to 74.5 AP, a net gain of +0.7 AP! This substantial increase robustly demonstrates that: KAGN's non-linear smoothing capability is the true core driver of the performance breakthrough. Furthermore, we explored a more aggressive KAGN-Fuse scheme (Stage 6) (illustrated in \cref{fig:kagn_fuse}), which reconfigured all fusion layers with KAGN. However, its performance (74.3 AP) did not surpass our final model (74.5 AP).

\section{Conclusion}
In this work, we set out to address the fine-grained feature loss and poor multi-scale adaptability of ViTPose, issues stemming from its overly simplistic Patch Embedding. Through a series of rigorous ablation studies, we revealed a critical insight: the model's performance bottleneck lies not in feature refinement, but in the fusion quality of the final multi-scale features.

Guided by this, we proposed the KAN-FPN-Stem, a novel hybrid architecture. It precisely addresses this bottleneck by replacing the FPN's terminal smoothing convolution with a KAGN convolution to adaptively rectify fusion artifacts. This ultimately achieved a significant +2.0 AP performance boost over the ViTPose-S baseline. This work not only validates that the Quality of multi-scale feature fusion is far more critical than feature Refinement, but also provides new insights for designing more efficient CNN-ViT hybrid networks.

\bibliographystyle{unsrt}
\bibliography{references}
\end{document}